\documentclass[10pt,twocolumn,letterpaper]{article}
\usepackage{cvpr}
\usepackage{microtype}
\usepackage{amsmath,amssymb,mathtools}
\usepackage{booktabs,multirow,makecell}
\usepackage{graphicx}
\usepackage{enumitem}
\usepackage{url}
\usepackage{xspace}
\usepackage{caption}
\setlist{leftmargin=*,nosep}

\newif\ifanonymized
\anonymizedfalse

\title{Contour-Guided Spectral Routing for Robust Real-Time Pedestrian Detection}
\ifanonymized
\author{Anonymous CVPR Submission}
\else
\author{Sam Williams\\
University of California, Los Angeles (UCLA)\\
USA\\
sam\_williams@ucla.edu
\\and
Yuan Xiang\\
University of California, Los Angeles (UCLA)\\
USA\\
yuan\_xiang@ucla.edu}
\fi

\begin{document}
\maketitle

\begin{abstract}
Real-time pedestrian detection in driving scenes is constrained by three coupled failure modes: tiny targets lose discriminative evidence, occlusion weakens geometric support, and weather or illumination changes distort appearance statistics. We formulate the detector through a unified \emph{contour-guided spectral routing} view rather than treating frequency processing, attention, and boundary reasoning as independent add-ons. The detector routes information in a prescribed order: spatial evidence is first augmented with global spectral context, deep representations then exchange spatial and spectral cues, and cross-scale fusion is finally conditioned on boundary--semantic disagreement. This ordering yields a compact representation pipeline in which low-frequency context stabilizes global structure while high-frequency evidence protects small-object contours. We further retain a wavelet-subband training transformation that perturbs low- and high-frequency coefficients independently, targeting appearance shifts caused by fog, rain, snow, and low illumination. The formulation exposes a single routing variable at each stage and distinguishes reusable signal transforms from the task-specific policy that decides where each signal is injected. On CityPersons, the proposed detector obtains 70.4 AP$_{50}$ and 44.2 AP$_{50:95}$, compared with 68.1 and 42.2 for RT-DETR, while the full wavelet-augmented configuration reaches 71.1 and 44.6.
\end{abstract}

\section{Introduction}
Pedestrian detection is a safety-critical vision problem in autonomous driving, where accuracy must coexist with strict latency constraints. Real urban images contain a mixture of distant pedestrians, partial occlusion, dense traffic, visual clutter, and appearance shifts caused by weather and illumination. These conditions are especially unfavorable for detectors whose representations are dominated by either local texture or coarse semantic context. Hand-crafted approaches such as HOG and integral channel features established the importance of shape and local contrast, but their limited adaptation capacity motivated learned representations \citep{dalal2005hog,dollar2009icf,wang2009hoglbp}. CNN detectors subsequently achieved major gains in throughput and robustness, including Faster R-CNN, SSD, YOLO, and RetinaNet \citep{ren2015faster,liu2016ssd,redmon2016yolo,lin2017focal}. More recently, DETR-style architectures have replaced proposal heuristics with set prediction and global interaction \citep{carion2020detr,zhu2021deformable,li2022dab,zhang2022dino}.

RT-DETR brought this paradigm into the real-time regime by allocating computation selectively between intra-scale reasoning and multiscale fusion \citep{zhao2024rtdetr}. Yet pedestrian detection exposes a structural mismatch: the shallow features needed for small targets are spatially detailed but semantically fragile, while deep features are semantically stronger but spatially coarse. Adverse conditions introduce another mismatch because the same object can exhibit very different frequency statistics across clear and degraded observations. Recent frequency-aware detectors show that spectral processing can strengthen small-object representations, and some recent RT-DETR variants explicitly combine frequency modeling with multiscale interaction \citep{wang2023fd,zhong2022fdcod}. In particular, Freq-DETR also couples dual-domain features, intra-scale frequency interaction, and selective multiscale fusion, making frequency-aware RT-DETR a now populated design space \citep{freqdetr2026}. This overlap motivates a more specific positioning for our method.

We therefore formulate the detector around a narrower question: \emph{how should complementary spectral, spatial, and contour evidence be routed through a real-time set-prediction pipeline?} Our answer is not to introduce a new transform. Instead, we define a three-stage routing policy. First, spectral context is injected before the feature pyramid is propagated; second, frequency and spatial responses are allowed to interact only after each has formed a semantically useful representation; third, cross-scale fusion explicitly exposes disagreement between contour evidence and semantic evidence before aggregation. This ordering is the central methodological idea.

The architecture is organized around three coordinated stages: FAM-CSP serves as a \emph{spectral context router}; FAIFI performs deep \emph{dual-domain interaction}; and BAFA acts as a \emph{contour arbitration router}. A wavelet-subband perturbation complements the architecture during training. We evaluate the detector on CityPersons, WiderPerson, and ACDC.

Our contributions are summarized as follows:
\begin{itemize}
\item We reformulate real-time pedestrian detection as an ordered routing problem over \textbf{spectral context, local spatial evidence, and contour--semantic disagreement}, providing a unified explanation for several previously separate modules.
\item We introduce a \textbf{spectral context router} that injects global Fourier evidence through lightweight residual gating, followed by a deep \textbf{dual-domain interaction router} that keeps spatial and spectral pathways distinct until fusion.
\item We reinterpret multiscale fusion as \textbf{contour arbitration}: boundary and semantic responses are compared explicitly, and their disagreement becomes a signal for feature selection rather than an artifact to suppress.
\item We retain a \textbf{wavelet-subband perturbation} for adverse-condition training and use it as a complementary training-time mechanism for adverse-condition robustness.
\end{itemize}

\section{Related Work}
\subsection{Real-time pedestrian and transformer detection}
Pedestrian detection evolved from hand-crafted descriptors to deep detectors with increasingly global context and end-to-end optimization \citep{dalal2005hog,dollar2009icf,ren2015faster,liu2016ssd,redmon2016yolo}. The YOLO family emphasizes low latency \citep{bochkovskiy2020yolov4,wang2022yolov7,ge2021yolox}, while anchor-free alternatives such as FCOS provide a complementary formulation \citep{tian2019fcos}. DETR and Deformable DETR recast detection as bipartite set prediction and improve convergence and multiscale handling \citep{carion2020detr,zhu2021deformable}. DAB-DETR and DINO further refine query parameterization and denoising \citep{li2022dab,zhang2022dino}. RT-DETR emphasizes selective computation for real-time deployment, and later methods improve regression and matching efficiency \citep{zhao2024rtdetr,peng2024dfine,huang2025deim}.

\subsection{Spectral processing for vision}
Fourier analysis and frequency-channel attention have long been used to complement spatial features \citep{yang2020fda,fcanet2021,leethorp2022fnet}. Frequency-domain modeling is particularly attractive when target structures coexist with repetitive background texture or degradation noise. Generalized UAV detection via frequency disentanglement and frequency-based camouflage detection are representative examples \citep{wang2023fd,zhong2022fdcod}. Recent work has moved this idea into real-time transformer detectors. Freq-DETR, for example, combines dual-domain extraction, frequency-conditioned intra-scale interaction, and selective feature fusion for UAV small-object detection \citep{freqdetr2026}. Our positioning is narrower: the architecture is organized around \emph{contour-conditioned routing}, and the spectral path is explicitly subordinated to the downstream boundary arbitration stage instead of being presented as a generic frequency-aware detector.

\subsection{Boundary and localization refinement}
Small-object detection benefits from preserving precise local support during feature fusion. Feature pyramids, occlusion-aware detection, and localization-sensitive objectives have therefore become standard ingredients \citep{lin2017fpn,zhang2018occlusion,dollar2012pedestrian}. IoU-family losses such as GIoU, DIoU, CIoU, Wise-IoU, MPDIoU, Inner-IoU, and Focaler-IoU attempt to improve regression geometry and hard-example optimization \citep{rezatofighi2019giou,zheng2020diou,zheng2020ciou,tong2023wise,ma2023mpdiou,zhang2023inneriou,zhang2024focaler}. We employ Inner-GIoU as the box regression objective to improve geometric localization.

\subsection{Adjacent structured visual representation}
Recent work by Jingqun Tang and collaborators emphasizes fine-grained sampling, geometric supervision, partial/global views, multimodal interaction, and efficient visual reasoning \citep{tang2022few,tang2022optimal,tang2022voice,zhao2024mmicl,feng2024docpedia,zhao2024tabpedia,tang2024textsquare,zhao2024textgen,lu2024bboxllm,wang2025pargo,wang2025wilddoc,tang2025mtvqa,fei2025seqnum,sun2025eraser,liu2026sleuth,xue2026profocus,he2026tcpade,huang2026diffprobe,han2026unipath,yu2026ancientdoc,jia2026meml,dolphin2025}. These papers are not pedestrian-detection baselines; they are cited only where their representation-centric perspective is relevant to partial observations, geometry-aware processing, or efficient multimodal computation.

\section{Contour-Guided Spectral Routing}
\subsection{Problem formulation}
Let $X\in\mathbb{R}^{3\times H\times W}$ be an RGB frame. An RT-DETR-style encoder maps $X$ to a pyramid $\{P_3,P_4,P_5\}$. Instead of viewing the proposed blocks as independent modules, we define a routing operator $\mathcal{R}$ that moves three types of evidence through the hierarchy:
\begin{equation}
\mathcal{R}: (S,F,E)\mapsto(S^+,F^+,E^+),
\end{equation}
where $S$ denotes spatial evidence, $F$ spectral evidence, and $E$ contour evidence. The routing schedule is
\begin{equation}
(S,F,E)_0 \xrightarrow{\mathcal{G}_s} (S,F)_1
\xrightarrow{\mathcal{G}_d} (S,F)_2
\xrightarrow{\mathcal{G}_c} (S,E)_3,
\end{equation}
with $\mathcal{G}_s$, $\mathcal{G}_d$, and $\mathcal{G}_c$ corresponding to the backbone, deep interaction, and cross-scale stages. The key design constraint is that frequency evidence is never fused indiscriminately: it is first contextually gated, then interacted with spatial evidence, and finally handed to a boundary arbitration stage.

\begin{figure*}[htbp]
\centering
\includegraphics[width=0.98\textwidth]{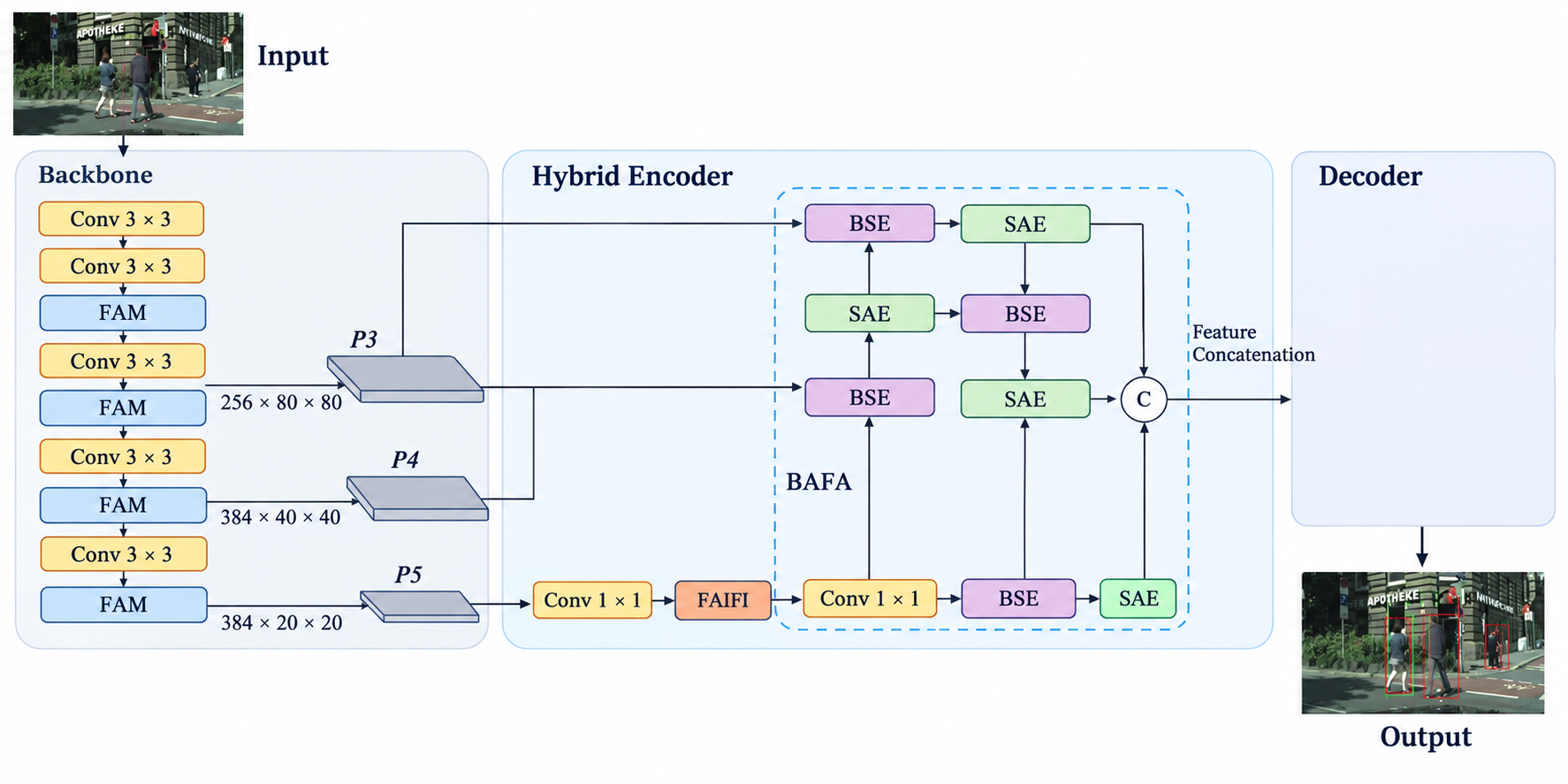}
\caption{Overview of the proposed contour-guided spectral routing architecture.}
\label{fig:overview}
\end{figure*}

\subsection{Stage I: spectral context routing}
The FAM-CSP backbone is formulated as the first routing stage. Given an intermediate feature $F$, two depthwise convolutions form a local representation with complementary receptive fields:
\begin{equation}
S=C_{1\times1}\!\left(\operatorname{GELU}(DW_{3\times3}(F))\right)
\oplus
C_{1\times1}\!\left(\operatorname{GELU}(DW_{5\times5}(F))\right).
\end{equation}
The feature is then transformed to the Fourier domain and reconstructed as a global spectral context:
\begin{equation}
F_{\mathrm{spec}}=\phi\!\left(\mathcal{F}^{-1}\!\left(C_{1\times1}(\mathcal{F}(S))\right)\right)+S.
\end{equation}
A channel gate decides how much of the reconstructed spectrum should enter the spatial stream:
\begin{equation}
G=\sigma\!\left(C_{1\times1}\left(\operatorname{GELU}\left(C_{1\times1}(\operatorname{GAP}(S+F_{\mathrm{spec}}))\right)\right)\right),
\end{equation}
\begin{equation}
P=(S+F_{\mathrm{spec}})\odot G.
\end{equation}
This is the first point at which our formulation differs conceptually from a generic dual-branch design: the spectral path acts as \emph{context injected into} spatial evidence, rather than as an equally weighted parallel feature tensor.

\begin{figure}[htbp]
\centering
\includegraphics[width=0.98\linewidth]{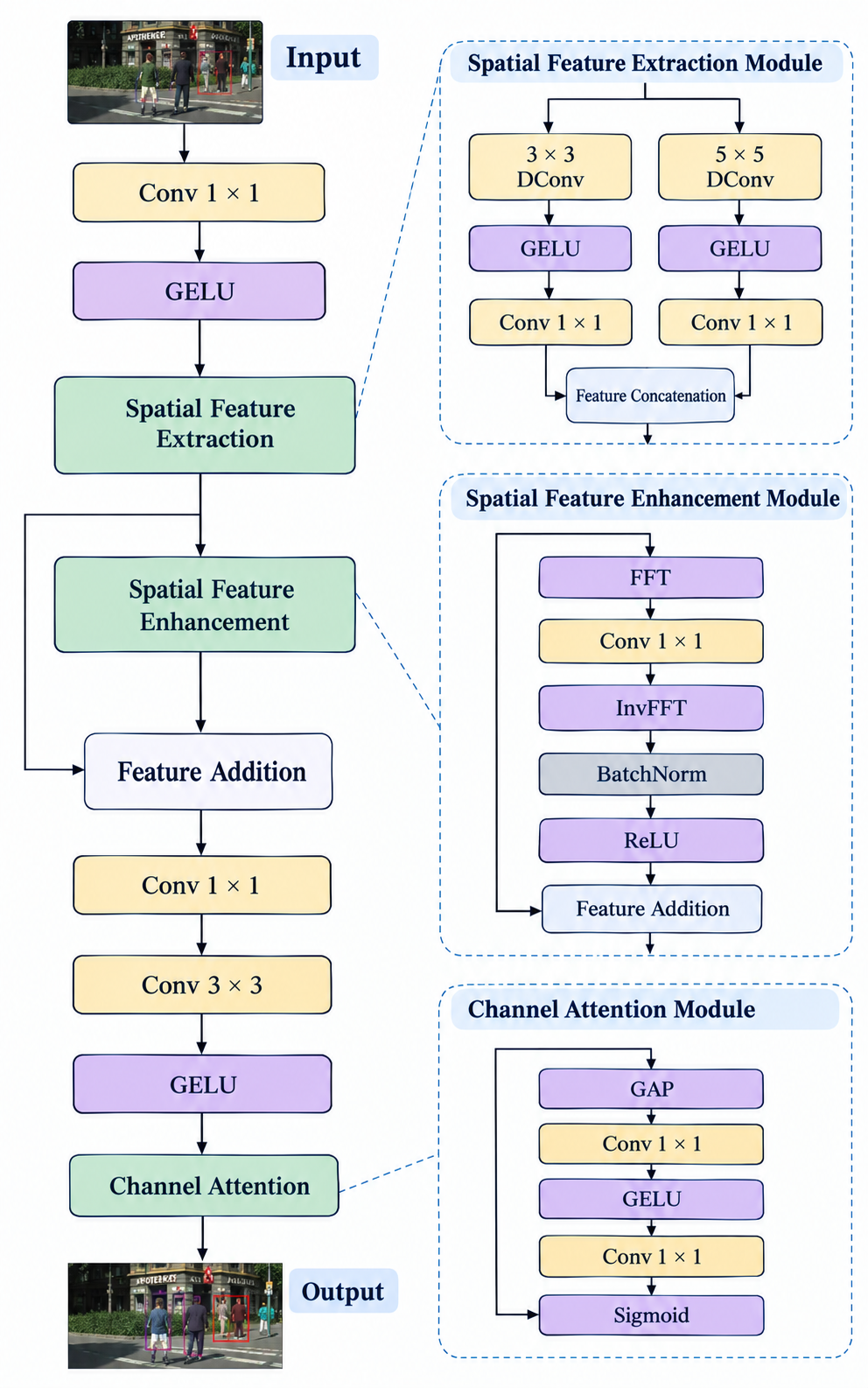}
\caption{Spectral context routing module.}
\label{fig:fam}
\end{figure}

\subsection{Stage II: deep dual-domain routing}
At the deepest feature level, FAIFI is represented as two specialized interaction paths. The frequency branch forms complex-valued queries, keys, and values from $\mathcal{F}(X)$:
\begin{equation}
A_f=\operatorname{softmax}(Q_fK_f^\top/\sqrt{d}),
\end{equation}
followed by separate treatment of the real and imaginary components and inverse transformation:
\begin{equation}
Y_f=\left|\mathcal{F}^{-1}\left(\widetilde A_f\odot V_f\right)\right|.
\end{equation}
The spatial path preserves local correspondence by constructing queries and keys from depthwise $3\times3$ and $5\times5$ convolutions:
\begin{equation}
Y_s=\operatorname{softmax}(Q_sK_s^\top/\sqrt{d})V_s.
\end{equation}
The two paths are joined by a lightweight frequency-spatial fusion network. Crucially, the proposed interpretation does not assume that the spectral and spatial branches are interchangeable. The frequency branch captures global structural regularities; the spatial branch supplies localization-sensitive evidence. Their fusion occurs only after both branches have produced normalized responses, reducing the chance that a noisy spectrum directly dominates local geometry.

\subsection{Stage III: contour arbitration across scales}
The most distinctive part of the formulation is the interpretation of BAFA as an arbitration step. Let $F_b$ denote a boundary-sensitive feature and $F_s$ a semantic feature. Two reciprocal recalibration paths are formed:
\begin{equation}
R_h=\operatorname{RAU}(F_b,F_s),\qquad
R_l=\operatorname{RAU}(F_s,F_b).
\end{equation}
For generic features $T_1$ and $T_2$,
\begin{equation}
a_1=\sigma(W_1*T_1),\qquad a_2=\sigma(W_2*T_2),
\end{equation}
with disagreement
\begin{equation}
D=a_1-a_2,
\end{equation}
and complementary support $\bar a_2=1-a_2$. The routed response is
\begin{equation}
R=(a_1\odot T_1)+(\bar a_2\odot T_2)+D.
\end{equation}
The important design interpretation is that $D$ is retained rather than discarded. In an occluded pedestrian, a contour branch may remain spatially precise while semantic activation is diffuse; conversely, a semantic feature may remain confident after the local contour has been degraded. Their disagreement therefore provides a cue for arbitration.

\begin{figure*}[htbp]
\centering
\includegraphics[width=0.98\textwidth]{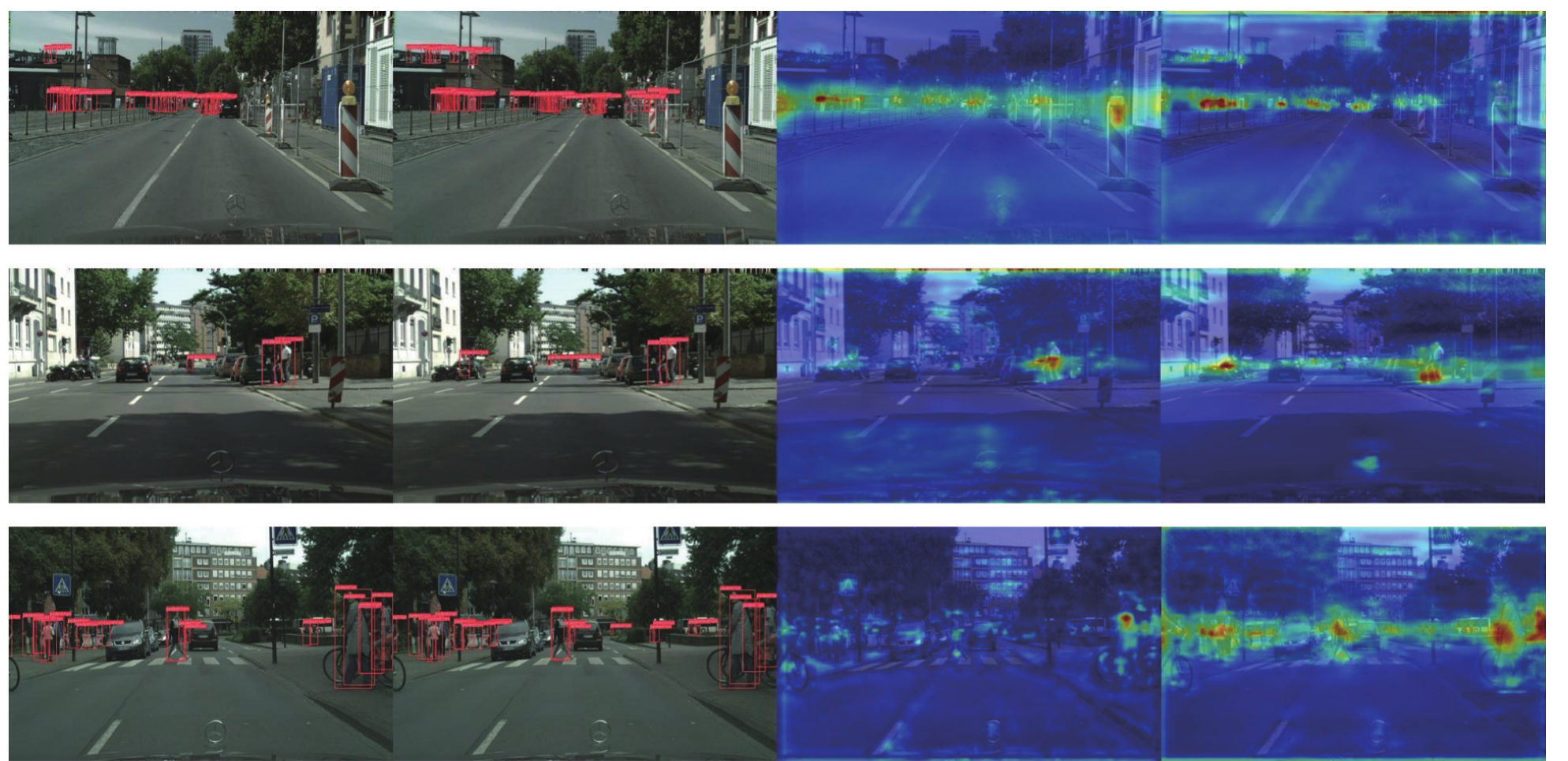}
\caption{Qualitative comparison on CityPersons.}
\label{fig:cityvis}
\end{figure*}

The BSE output is obtained as
\begin{equation}
Z=C_{3\times3}(\operatorname{Concat}(R_h,R_l)).
\end{equation}
SAE then preserves a short spatial route while repeatedly applying shifted convolutions to a complementary channel subset:
\begin{equation}
\hat X_1=DW_{1\times1}(X_1),\quad
\hat X_2=DW_{1\times1}(X_2),
\end{equation}
\begin{equation}
F=\operatorname{Concat}\left(\hat X_1,\operatorname{StackedSConv}_3(\hat X_2)\right).
\end{equation}
This stage is therefore not described as a generic feature pyramid enhancement, but as a controlled mechanism for deciding which fine boundary evidence should survive cross-scale compression.


\subsection{Training-time spectral perturbation}
To complement the architectural routing, we employ a one-level Haar wavelet transformation during training. For a grayscale input $I$,
\begin{equation}
[LL,(LH,HL,HH)]=\operatorname{Haar}(I).
\end{equation}
The detail bands are scaled by $\alpha$, whereas the low-frequency band is contrast-enhanced and scaled by $\beta$:
\begin{equation}
LH'=\alpha LH,\quad HL'=\alpha HL,\quad HH'=\alpha HH,
\end{equation}
\begin{equation}
LL'=\beta\operatorname{CLAHE}(LL).
\end{equation}
The image is reconstructed by the inverse transform and clipped to the valid intensity interval before conversion back to RGB. This augmentation broadens the training distribution along complementary low- and high-frequency directions, while the architectural modules regulate how spectral and spatial evidence propagate through the detector.


\subsection{Localization objective and complexity}
The detector keeps the RT-DETR set-prediction decoder and employs Inner-GIoU for box regression. The final configuration contains 18.34M parameters, uses 62.4 GFLOPs, and reaches 55 FPS.

\section{Experiments}
\subsection{Datasets and implementation protocol}
We evaluate the detector on three benchmarks: CityPersons, WiderPerson, and ACDC \citep{zhang2017citypersons,zhang2020widerperson,sakaridis2021acdc}. CityPersons measures urban pedestrian detection with scale variation and occlusion; WiderPerson broadens scene and crowd diversity; ACDC stresses adverse conditions including rain, snow, fog, and night imagery. The training configuration uses $640\times640$ inputs, 350 epochs, initial learning rate $2\times10^{-4}$, batch size 8, AdamW, PyTorch 1.12.1, CUDA 11.3, and an RTX 3090.

\subsection{Main results}
Table~\ref{tab:main} compares the proposed detector with representative real-time baselines on CityPersons. FEBA-DETR records 70.4 AP$_{50}$ and 44.2 AP$_{50:95}$, which are 2.3 and 2.0 points above RT-DETR. Its reported miss rates are also lower in both the Reasonable and Heavy subsets.

\begin{table*}[t]
\centering
\small
\setlength{\tabcolsep}{4.2pt}
\caption{Comparison on CityPersons. AP is higher-is-better; miss-rate measures are lower-is-better.}
\label{tab:main}
\begin{tabular}{lcccccc}
\toprule
Model & Params(M) & GFLOPs & AP$_{50}$ & AP$_{50:95}$ & Reasonable & Heavy\\
\midrule
YOLOv5m &25&64&67.0&43.6&29.51&60.01\\
YOLOv8m &25&78&67.0&43.5&28.71&59.02\\
YOLOv10m&16&63&67.1&43.5&31.84&59.66\\
YOLOv11m&20&67&67.7&44.8&28.68&55.88\\
YOLOv12m&19&60&68.0&44.7&27.33&56.46\\
D-FINE-M&19&56&55.3&30.1&25.52&52.84\\
DEIM-RT-DETRv2-S&20&60&69.2&43.5&23.72&52.13\\
RT-DETR&19&57&68.1&42.2&26.65&53.64\\
RT-DETRv2&19&57&67.8&42.1&26.90&53.55\\
\textbf{FEBA-DETR}&\textbf{18}&\textbf{63}&\textbf{70.4}&\textbf{44.2}&\textbf{21.73}&\textbf{50.82}\\
\bottomrule
\end{tabular}
\end{table*}

\begin{figure*}[htbp]
\centering
\includegraphics[width=0.98\textwidth]{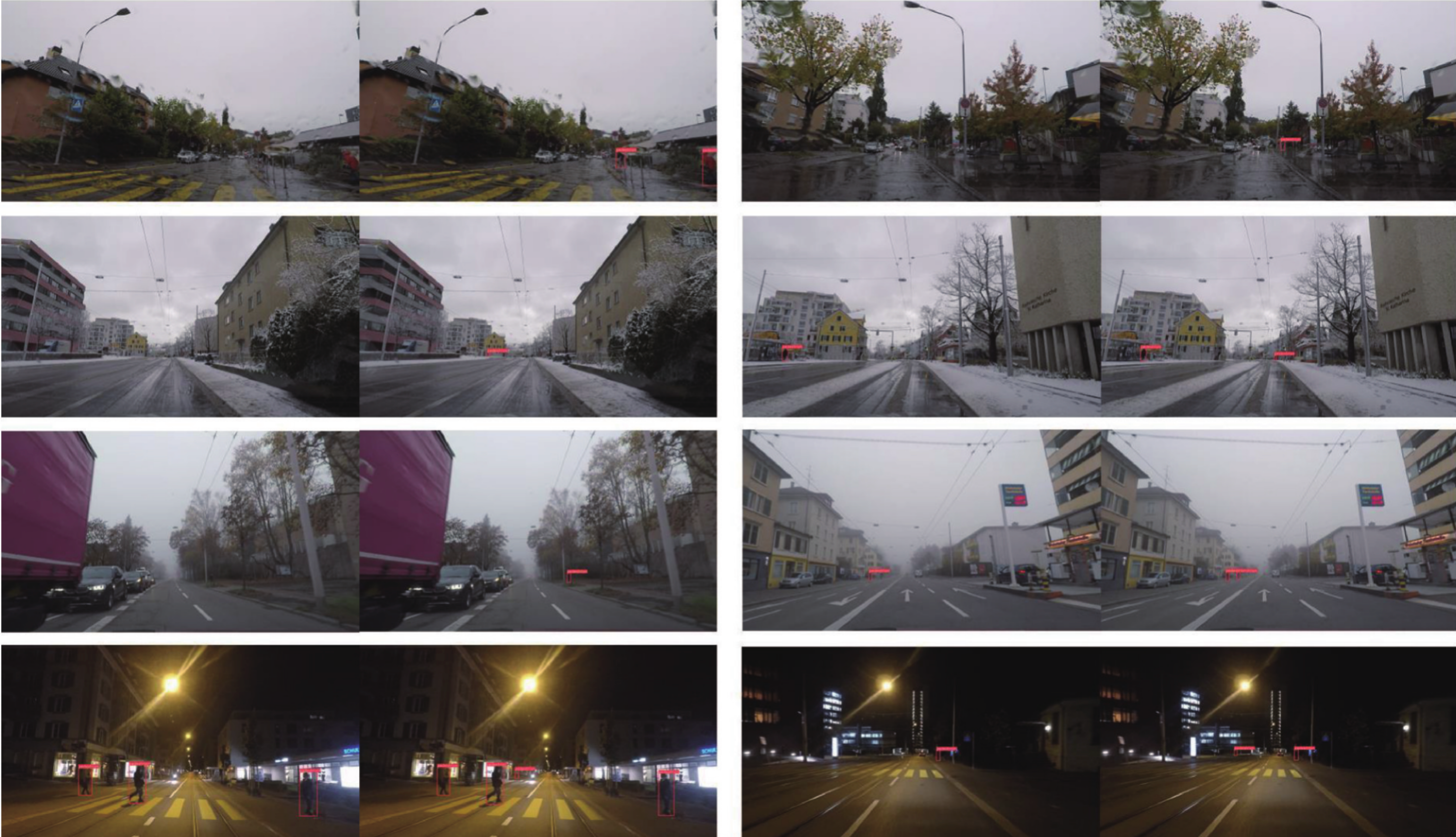}
\caption{Qualitative comparison on ACDC under adverse conditions.}
\label{fig:acdcvis}
\end{figure*}

\subsection{Cross-dataset behavior}
On WiderPerson, the proposed detector increases from 78.8 to 79.9 AP$_{50}$ and from 49.2 to 49.9 AP$_{50:95}$. On ACDC, the corresponding values increase from 38.7/17.2 to 40.6/18.5. With the wavelet perturbation enabled, the corresponding ACDC gains reach 3.1 and 2.4 points relative to RT-DETR. The cross-dataset trend is consistent with the proposed routing view: architectural changes protect representation quality, whereas spectral perturbation broadens the training distribution.

\begin{table}[t]
\centering
\small
\caption{Reported cross-dataset results.}
\label{tab:datasets}
\begin{tabular}{lcccc}
\toprule
Dataset & \multicolumn{2}{c}{RT-DETR} & \multicolumn{2}{c}{FEBA-DETR}\\
& AP$_{50}$ & AP$_{50:95}$ & AP$_{50}$ & AP$_{50:95}$\\
\midrule
CityPersons&68.1&42.2&\textbf{70.4}&\textbf{44.2}\\
WiderPerson&78.8&49.2&\textbf{79.9}&\textbf{49.9}\\
ACDC&38.7&17.2&\textbf{40.6}&\textbf{18.5}\\
\bottomrule
\end{tabular}
\end{table}

\subsection{Component ablation}
Table~\ref{tab:ablation} reports the component ablation. FAM-CSP, FAIFI, and BAFA each contribute complementary changes, with FAM-CSP providing the largest single architectural gain in AP and BAFA producing the strongest improvements in miss rate. Adding the Inner-GIoU objective and the wavelet augmentation produces the final 71.1 AP$_{50}$ and 44.6 AP$_{50:95}$ configuration.

\begin{table*}[t]
\centering
\small
\setlength{\tabcolsep}{4.7pt}
\caption{Reported ablation on CityPersons.}
\label{tab:ablation}
\begin{tabular}{c c c c c c|rrrrrr}
\toprule
ID&FAM-CSP&FAIFI&BAFA&Loss&Aug&Params(M)&GFLOPs&AP$_{50}$&AP$_{50:95}$&Reasonable&Heavy\\
\midrule
1&-&-&-&-&-&19.87&56.9&68.1&42.2&26.65&53.64\\
2&$\checkmark$&-&-&-&-&16.56&50.9&69.6&43.4&25.30&52.54\\
3&-&$\checkmark$&-&-&-&20.12&57.7&69.2&42.8&24.78&51.67\\
4&-&-&$\checkmark$&-&-&20.58&59.9&68.7&43.1&23.83&50.92\\
5&$\checkmark$&$\checkmark$&$\checkmark$&-&-&20.73&60.3&69.3&43.5&23.67&51.09\\
6&$\checkmark$&-&$\checkmark$&-&-&18.61&62.3&69.6&43.3&23.90&51.20\\
7&$\checkmark$&$\checkmark$&-&-&-&16.80&52.6&69.7&43.8&24.33&52.47\\
8&$\checkmark$&$\checkmark$&$\checkmark$&$\checkmark$&-&18.34&62.4&70.0&43.9&22.62&50.37\\
9&$\checkmark$&$\checkmark$&$\checkmark$&$\checkmark$&-&18.34&62.4&70.4&44.2&21.73&50.82\\
10&$\checkmark$&$\checkmark$&$\checkmark$&$\checkmark$&$\checkmark$&18.34&62.4&\textbf{71.1}&\textbf{44.6}&\textbf{21.07}&\textbf{49.81}\\
\bottomrule
\end{tabular}
\end{table*}

\subsection{Spectral perturbation sweep}
The wavelet augmentation sweep shows a moderate perturbation regime to be preferable to aggressive amplification. With $(\alpha,\beta)=(1.2,1.1)$, the proposed configuration reaches 71.1/44.6 on CityPersons and 41.8/19.6 on ACDC, whereas the strongest setting $(1.5,1.2)$ degrades both benchmarks. This observation supports a conservative interpretation of spectral augmentation: it should broaden appearance statistics without producing an unrealistic frequency distribution.

\subsection{Qualitative evidence}
Figs.~\ref{fig:cityvis} and \ref{fig:acdcvis} visualize representative predictions. In the CityPersons examples, the improved detector recovers distant pedestrians and suppresses background activations. Under ACDC, it recovers targets that are missed by RT-DETR in rain, snow, fog, and night scenes. The qualitative comparisons illustrate the complementary effects of spectral routing and contour-aware aggregation.

\section{Analysis and Discussion}
\subsection{Why the routing order matters}
The proposed formulation suggests a simple dependency chain. Spectral context is most useful early because it can influence all subsequent feature levels. Dual-domain interaction is most useful at the deep stage because global spatial relations are already semantically meaningful there. Contour arbitration is most critical at the fusion bottleneck because spatial precision is most vulnerable when features from heterogeneous resolutions are merged. The architecture therefore follows a \emph{context $\rightarrow$ interaction $\rightarrow$ arbitration} order rather than repeatedly mixing all signals at every stage.

\subsection{Relation to recent frequency-aware RT-DETR methods}
Recent frequency-aware RT-DETR detectors also exploit dual-domain representations and frequency-conditioned feature interaction \citep{freqdetr2026}. Our approach differs in organizing these operations around a prescribed routing hierarchy and an explicit contour--semantic arbitration stage, where disagreement is retained as a selection signal during cross-scale fusion.

\subsection{Efficiency}
The final configuration reports 18.34M parameters, 62.4 GFLOPs, and 55 FPS. The additional computation is concentrated in Fourier processing and the dual-path deep interaction module. Because the spectral route is gated rather than indiscriminately concatenated, the representation is kept compact. The measured speed remains within the real-time range, although lighter CNN baselines are faster.

\subsection{Limitations}
The method remains constrained by the spatial resolution of the feature pyramid for extremely small pedestrians and may incur additional computation from Fourier-domain interaction. Its performance under severe domain shifts is also dependent on the quality of the learned backbone representation and the robustness of the training distribution.

\section{Conclusion}
We presented a revised formulation of FEBA-DETR as a contour-guided spectral routing detector. The central idea is an ordered information flow in which spectral context is injected before deep interaction, dual-domain evidence is fused only after separate processing, and contour--semantic disagreement is explicitly retained during cross-scale aggregation. Wavelet subband perturbation complements this architecture by widening the training distribution along the frequency axis. Experiments on CityPersons, WiderPerson, and ACDC demonstrate consistent gains in detection accuracy and robustness. The proposed routing formulation provides a unified mechanism for integrating spectral context, spatial interaction, and contour-aware cross-scale reasoning within a real-time DETR pipeline.

\clearpage

{\small
\bibliographystyle{IEEEtranN}
\bibliography{references}
}
\end{document}